\documentclass{article}

\usepackage[nonatbib,preprint]{neurips_2024}

\usepackage[utf8]{inputenc} % allow utf-8 input
\usepackage[T1]{fontenc}    % use 8-bit T1 fonts
\usepackage{hyperref}       % hyperlinks
\usepackage{url}            % simple URL typesetting
\usepackage{booktabs}       % professional-quality tables
\usepackage{amsfonts}       % blackboard math symbols
\usepackage{nicefrac}       % compact symbols for 1/2, etc.
\usepackage{microtype}      % microtypography
\usepackage{xcolor}         % colors
\usepackage{graphicx}
\usepackage{amsmath}
\usepackage{amsthm}
\usepackage{amssymb}
\usepackage{dsfont}
\usepackage{wrapfig}
\usepackage[numbers]{natbib}

\title{Understanding Visual Concepts Across Models}

\author{%
  Brandon Trabucco$^{1}$, Max Gurinas$^{2}$, Kyle Doherty$^{3}$, Ruslan Salakhutdinov$^{1}$ \\
  $^{1}$Carnegie Mellon University, $^{2}$University Of Chicago Laboratory Schools, $^{3}$MPG Ranch \\
  \texttt{brandon@btrabucco.com}, \texttt{rsalakhu@cs.cmu.edu}
}

\begin{document}

\maketitle

\begin{abstract}
Large multimodal models such as Stable Diffusion can generate, detect, and classify new visual concepts after fine-tuning just a single word embedding. Do models learn similar words for the same concepts (i.e. \texttt{<orange-cat>} = orange + cat)? We conduct a large-scale analysis on three state-of-the-art models in text-to-image generation, open-set object detection, and zero-shot classification, and find that new word embeddings are model-specific and non-transferable. Across 4,800 new embeddings trained for 40 diverse visual concepts on four standard datasets, we find perturbations within an $\epsilon$-ball to any prior embedding that generate, detect, and classify an arbitrary concept. When these new embeddings are spliced into new models, fine-tuning that targets the original model is lost. We show popular soft prompt-tuning approaches find these perturbative solutions when applied to visual concept learning tasks, and  embeddings for visual concepts are not transferable.  Code for reproducing our work is available at: \href{https://visual-words.github.io}{visual-words.github.io}.
\end{abstract}

\section{Introduction}
\label{sec:intro}

Fine-tuning prompts is a widely successful technique for adapting large pretrained models to new tasks from limited data \cite{prompt-tuning,prefix-tuning,auto-prompt,textual-inversion}. In language modelling, these prompts can efficiently teach pretrained language models specialized tasks, such as reading tables \cite{prefix-tuning}. In text-to-image generation, they can embed subjects with unique, often hard-to-describe appearances into the generations of a diffusion model \cite{textual-inversion,dreambooth}. Large multimodal models, such as Stable Diffusion \cite{stable-diffusion}, OWL-v2 \cite{owl-vit}, and CLIP \cite{clip,open-clip,dfn}, can generate, detect, and classify diverse visual concepts not present in their training data after fine-tuning just a single word embedding representing that concept in their prompt \cite{da-fusion}. \textit{Do these models learn similar words for the same visual concept?} There is an emerging hypothesis in multimodal machine learning that text-based models learn to process visual information \cite{llava,fromage}, and acquire similar representations for visual information \cite{platonic-representation,universal-computation}, despite training purely on text. This investigation aims to determine if the hypothesis extends to soft prompts that encode specific visual concepts, and whether these prompts converge to a solution that different models can interpret.

For example, do text-based models that can generate, detect, and classify various species of cats learn similar words for orange cats (i.e. \texttt{<orange-cat>} = orange + cat)? We conduct a large-scale analysis on three state-of-the-art models in text-to-image generation, open-set object detection, and zero-shot classification, and find that new word embeddings are model-specific and non-transferable. We optimize 4,800 new embeddings for Stable Diffusion \cite{stable-diffusion}, OWL-v2 \cite{owl-vit}, and CLIP \cite{clip,open-clip,dfn} to generate, detect, and classify 40 diverse visual concepts in four standard datasets with high fidelity. Interestingly, \texttt{<orange-cat>} $\neq$ orange + cat for any model and concept tested. Instead, for all tested models we find perturbations within an $\epsilon$-ball to any prior embedding that generate, detect, and classify an arbitrary visual concept. We refer to this behavior as \textbf{fracturing} of the embedding space. Fractured models have several noteworthy properties. First, their prompts are difficult to interpret: prompts for orange cats may be close to prompts for blue cars, and far from prompts for black cats. Second, their prompts are not transferable: when embeddings trained for one model are spliced into the prompt of a new model, the second model ignores fine-tuning that targets the original model.

\begin{figure}[t]
    \centering
    \includegraphics[width=\linewidth]{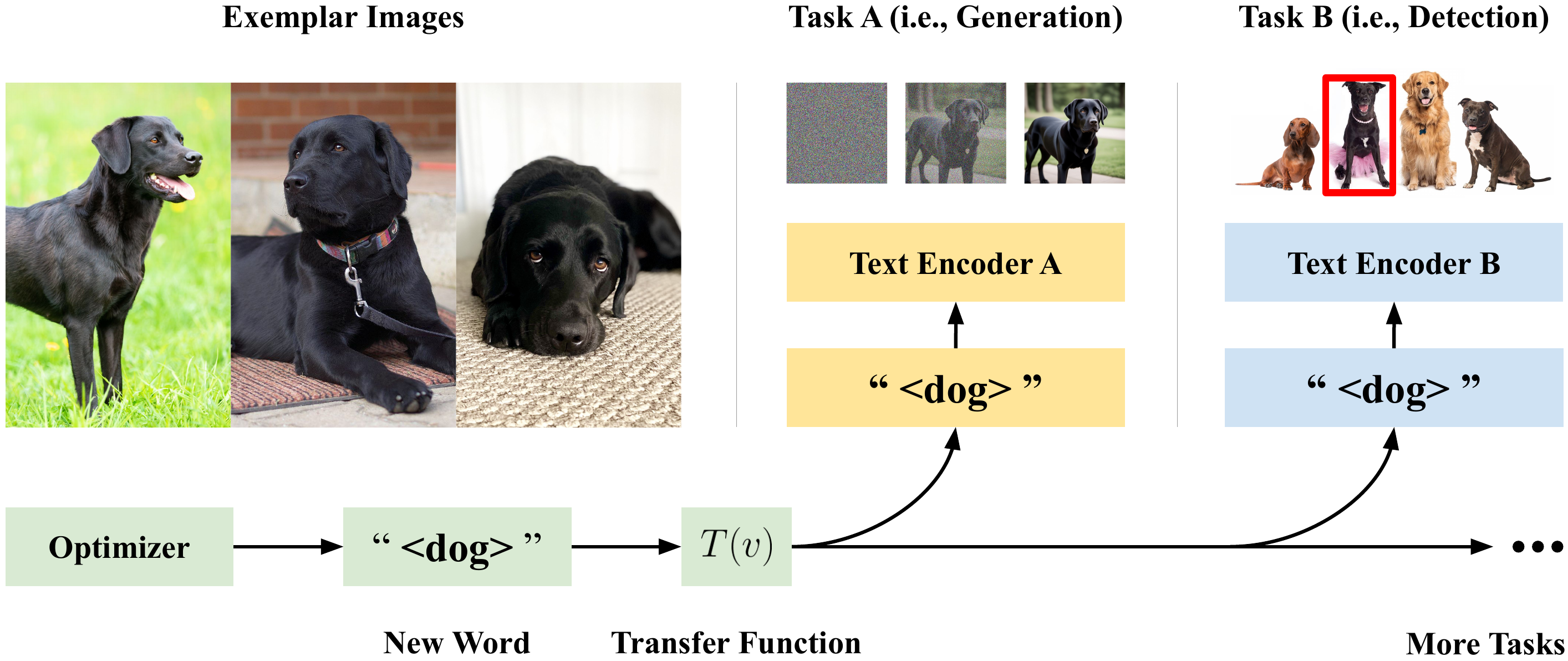}
    \vspace{0.2cm}
    \caption{\small Large multimodal models can learn new words that represent specific concepts, like \texttt{<black-dog>} for the black Labrador retriever on the left in the figure. Do models learn similar words for the same concept? We study the interoperability of new word embeddings that encode visual concepts across three models and tasks, and show that popular soft prompt-tuning approaches find model-specific and non-transferable solutions.}
    \label{fig:teaser}
\end{figure}

Popular soft prompt-tuning approaches \cite{textual-inversion} find solutions that are model-specific and non-transferable when applied to visual concept learning tasks. Initialization does not matter, and solutions found near words unrelated to the visual concept perform just as well as related words. When embeddings are transferred using a linear map to a second model, illustrated in Figure~\ref{fig:teaser}, embeddings snap to the nearest word in the embedding space known to both models, and all fine-grain optimization is lost. This suggests new embeddings that encode visual concepts cannot be expressed via simple algebraic combinations of the existing word embeddings  (i.e. \texttt{<orange-cat>} $\neq$ orange + cat). Instead, these new embeddings resemble perturbations akin to adversarial examples \cite{adversarial-examples,adversarial-patch}, and the perturbations that generate, detect, and classify specific concepts like "orange cat" are different for each model.

This work contributes a large-scale study of soft prompts that encode specific visual concepts across generation, detection, and classification tasks. We provide a benchmark for training such prompts on a diverse set of visual concepts, and evaluating their transferability across three models. Our work aims to galvanize the interoperability of large multimodal models following Figure~\ref{fig:teaser}, allowing prompts trained for generating black Labradors to be re-used for detection, and other tasks. Transferring prompts can significantly improve the adaptability and cost of machine learning systems by eliminating the need to re-train prompts when new models are released. We highlight the difficulty of transferring prompts for current models, investigate what these prompts learn, and what happens when they are transferred. Code for reproducing our work, and benchmarking new transfer methods is available at the following official website: \href{https://visual-words.github.io}{visual-words.github.io}.

\section{Related Works}
\label{sec:related-works}

\paragraph{Text-To-Image Generation.} With the advent of diffusion-based architectures, large-scale generative models have developed impressive photo-realism. Approaches like Stable Diffusion \cite{stable-diffusion}, DALL-E 2 \cite{dalle2}, and Imagen \cite{imagen} employ diffusion-based approaches \cite{ddpm, diffusion} that start from an initial Gaussian noise map, and iteratively denoise the image over several denoising diffusion steps. These approaches incorporate pretrained text-encoders, such as CLIP \cite{clip} in Stable Diffusion \cite{stable-diffusion}, to guide generation in the diffusion process. Guidance is typically applied through Classifier-free Guidance \cite{classifier-free-guidance}, which allows the influence of the text-encoder to be increased, at the expense of generation quality. Diffusion models have remarkable flexibility, and can generate new subjects from a handful of examples by learning embeddings for pseudo tokens representing the subject in the prompt \cite{textual-inversion, da-fusion}. Fine-tuning both the model and the prompt, as in Dreambooth \cite{dreambooth}, leads to improved generation of subjects, while retaining the controllability of pseudo tokens. These pseudo tokens for diffusion models are an instance of prompts encoding visual concepts, and our analysis applies to them.

\paragraph{Open-Vocabulary Object Detection.} Parallel to work in generation, large-scale object detection models have developed a comparable strong versatility, and can detect new objects from short descriptions of their appearance (i.e. detect \textit{black dog})  \cite{seem,grounding-dino,owl-vit}. Models like Grounding DINO \cite{grounding-dino}, SEEM \cite{seem}, and OWLv2 \cite{owl-vit} employ a pretrained text encoder to produce representations for classifying bounding boxes. In OWLv2 \cite{owl-vit}, representations from a pretrained CLIP \cite{clip} text encoder are contrasted with region-based representations from a vision transformer backbone. Grounding DINO \cite{grounding-dino}, and SEEM \cite{seem} employ representations from a pretrained text encoder (BERT \cite{bert}, and UniCL \cite{unicl}, respectively) to directly guide bounding box proposal. We show open-vocabulary object detectors can detect new objects from a handful of examples by optimizing a single new word embedding for the object in their prompt. Furthermore, our analysis shows many of the properties of these new words in open-vocabulary object detection are the same as for text-to-image generation.

\paragraph{Zero-Shot Classification.} We use CLIP \cite{clip} for zero-shot classification. We insert new word embeddings optimized for classifying new visual concepts in the prompt of the CLIP text encoder, and contrast text representations with image representations from the CLIP vision encoder on test images. Prior work shows CLIP is an effective zero-shot classifier on open-vocabulary tasks \cite{clip,chils}. We use checkpoints from OpenAI CLIP \cite{clip}, OpenCLIP \cite{open-clip}, and Data Filtering Networks \cite{dfn}, trained on LAION-5B \cite{laion}. Diffusion models can also be used as zero-shot classifiers \cite{diffusion-classifier-cmu, diffusion-classifier-berkeley}, but we focus on CLIP for better task coverage. Our analysis shows that soft prompts learned for zero-shot classification share properties with open-vocabulary object detectors and text-to-image models.

\paragraph{Prompt-Tuning.} The word embeddings we optimize for visual concept learning tasks are closely related to prompt-tuning \cite{prompt-tuning, prefix-tuning}. Prompt tuning aims to find a prefix or an entire prompt that causes a pretrained language model to perform a specialized task, such as reading tables \cite{prompt-tuning, prefix-tuning}. These methods treat the prompt as a trainable parameter, and optimize the embeddings of the prompt to minimize a task loss function. Prior work has shown the resulting soft prompts in language modelling tasks are hard to interpret \cite{prompt-waywardness}, as their closest discrete prompts are often unrelated to the desired task. Transferring learned prompts is an important task in jail-breaking LLMs \cite{gcg,smooth-llm}, and researchers are searching over discrete prompts \cite{hard-prompts-made-easy, auto-prompt, gcg, smooth-llm}. In pure language modelling tasks, researchers have shown that certain soft prompts can transfer between models with the same architecture and task, but different weights \cite{continuous-prompt-generation,combination-of-discrete-prompts,zero-shot-prompt-transfer}. We extend this investigation to visual tasks, and models with different architectures, trained on different label modalities (images, bounding boxes, and class labels).

\paragraph{Adversarial Examples.} The perturbative structure of word embeddings that encode visual concepts are akin to an adversarial attack on the embeddings of text encoders. Adversarial robustness is an extensively studied field in computer vision \cite{adversarial-examples}, with a variety of attack methods, including \cite{adversarial-examples, adversarial-patch, pgd, bim, cw}, and defense methods, including \cite{pgd, local-linearization, ensemble-training, adv-logit-pairing}. Adversarial attacks in computer vision traditionally focus on modifying the pixels in an image, whereas we modify word embeddings. Adversarial attacks on language are in their infancy, including jail-breaking approaches \cite{gcg,smooth-llm}, and typically involve searching over discrete prompts \cite{lm-attack-survey,bert-attack}, rather than continuous embeddings.

\section{Transfer Evaluation Methodology}
\label{sec:transferring-visual-words}

Finding words for visual concepts across models involves finding a map between the embedding spaces of different models. We call this mapping the Transfer Function $T(v)$, depicted in Figure~\ref{fig:teaser}. The goal of the Transfer Function is to map word representations for visual concepts from the vector space $\mathcal{X} = \mathbb{R}^{d_{x}}$ for word embeddings in one model, to the vector space $\mathcal{Y} = \mathbb{R}^{d_{y}}$ for word embeddings in another model. $\mathcal{X}$ may correspond to Stable Diffusion \cite{stable-diffusion} word embeddings for a generation task, and $\mathcal{Y}$ may be OWL-v2 \cite{owl-vit} word embeddings for a detection task. Given these vector spaces, the Transfer Function predicts the representation $\vec{x}(w)$ in the vector space $\mathcal{X}$ for a word $w$ originally from the vector space $\mathcal{Y}$ given just the word vector representation $\vec{y}(w)$.
\begin{equation}\label{eqn:transfer-definition}
    T^{\;y \to x} : \mathcal{Y} \to \mathcal{X} = \arg\min_{T} \; \mathbb{E}_{w \sim p_{w}} \| \vec{x}(w) - T(\vec{y}(w)) \|^2_2
\end{equation}

The Transfer Function $T(v)$ minimizes the average prediction error between transferred word embeddings $T(\vec{y}(w))$ and real word embeddings $\vec{x}(w)$ from the vector space $\mathcal{X}$. We average this prediction error over a uniform distribution $p_{w}$ of the words that exist in both vector spaces $\mathcal{X}$ and $\mathcal{Y}$. In our experiments on Stable Diffusion 2.1 \cite{stable-diffusion} and OWL-v2 \cite{owl-vit}, the number of words in $p_{w}$ is large (> 40,000), much larger than the number of components $d_{x}$ and $d_{y}$ in each vector space. 
%This large amount of paired examples encourages learning a deep neural network, but parameterizing $T(v)$ is a challenging question on its own. For simplicity we focus on linear functions, leaving other choices as future work.

\subsection{Finding A Linear Transfer Function}

Solving the optimization problem given by Equation~\ref{eqn:transfer-definition} is hard in general, and to simplify the investigation, we restrict our focus to the class of linear Transfer Functions. This restriction transforms the hard problem in Equation~\ref{eqn:transfer-definition} into a Linear Least Squares estimator, which has a closed-form solution. Consider a pair of matrices $X \in \mathbb{R}^{n \times d_{x}}$ and $Y \in \mathbb{R}^{n \times d_{y}}$, where each pair of rows in $X$ and $Y$ is a pair of word vector embeddings $\vec{x}(w)$ and $\vec{y}(w)$ for a word $w$ contained in the support of the distribution $p_{w}$. The Linear Least Squares estimator we employ for $T^{\;y \to x}$ is given below.
\begin{equation}\label{eqn:linear-transfer-function}
    T^{\;y \to x} = \arg\min_{T} \; \mathbb{E}_{w \sim p_{w}} \| \vec{x}(w) - T \vec{y}(w) \|^2_2 = ( Y^{T} Y )^{-1} Y^{T} X
\end{equation}

One can interpret the map $T^{\;y \to x}$ as lining up the directions in the vector spaces $\mathcal{X}$ and $\mathcal{Y}$ that correspond to the same visual concepts. Word embeddings often have algebraic relationships~\cite{word2vec}, and a linear Transfer Function preserves these relationships by distributing over addition.

\begin{figure}[t]
    \centering
    \includegraphics[width=\linewidth]{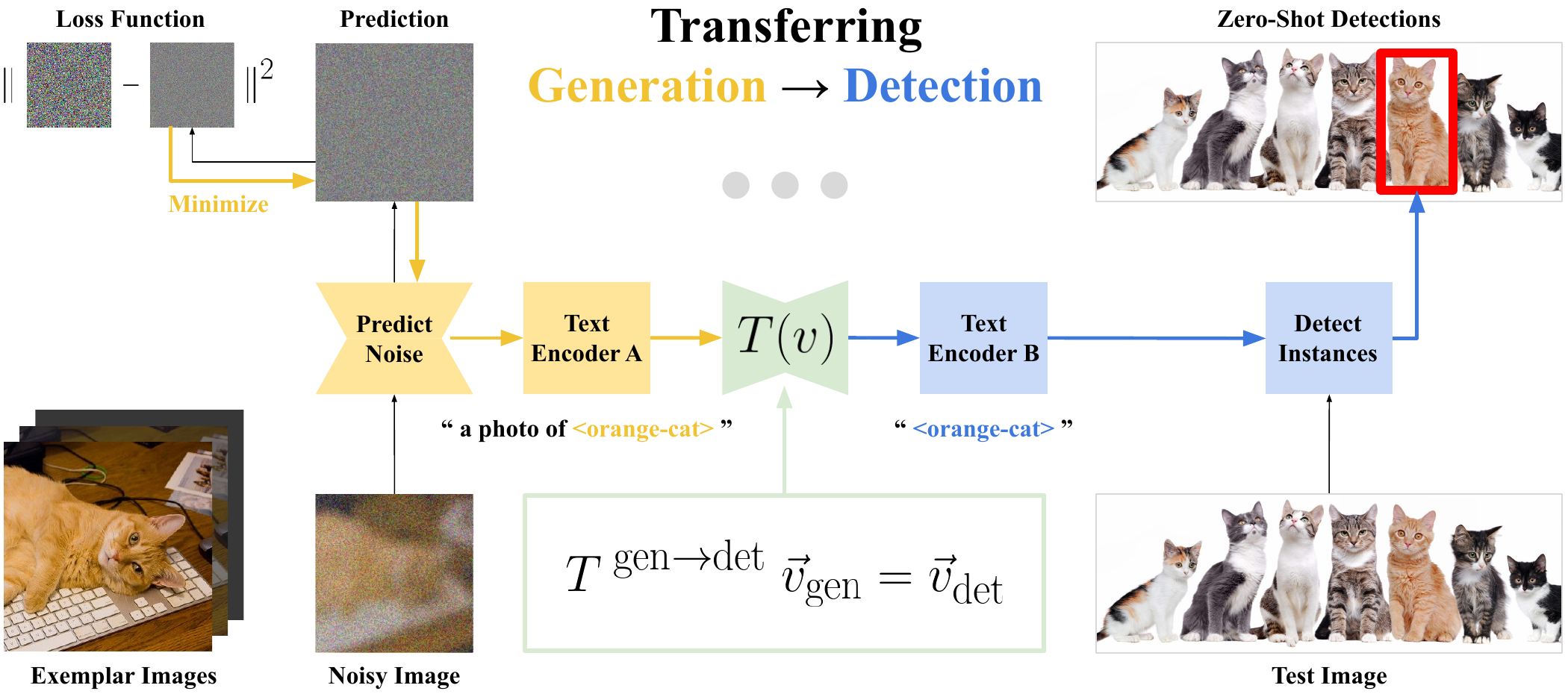}
    \caption{\small Transferring words optimized for generation to detection tasks. We fine-tune the vector embeddings for new words (such as \texttt{<orange-cat>} for the orange cat in the figure) to minimize a noise prediction loss for generation. Vector embeddings are transferred from generation to detection using the Transfer Function $T(\vec{v})$, and used to produce zero-shot instance detections for the target visual concept (in this case, orange cats).}
    \label{fig:transfer-generation}
\end{figure}

%RUSS: Can we avoid using foundation model in the figure -- OWL is not a foundatiuon model... 
% Brandon: yes, will do

\subsection{Evaluating Words On Transferred Tasks}

Using Equation~\ref{eqn:linear-transfer-function}, we estimate Transfer Functions between all six ordered subsets of three state-of-the-art models, and evaluate words optimized for visual concepts on one task (such as generation), and transferred to the same visual concepts on another task (such as classification). Consider a dataset $D$ of images $I$ depicting a specific visual concept, such as a black Labrador retriever, and task-specific annotations $a_{y}$, such as bounding boxes ($a_{y} \in \mathbb{R}^{b \times 4}$), or class labels ($a_{y} \in \mathbb{N}$). We first optimize word vector embeddings $\vec{v}_{y} \in \mathcal{Y}$ to minimize a task-specific loss function $\mathcal{L}_{y}$. We then zero-shot transfer $\vec{v}_{y}$ to task $x$ using the linear map $\vec{v}_{x} = T^{\;y \to x} \vec{v}_{y}$, and evaluate a task-specific performance metric $\mathcal{M}_{x}$. Loss functions and performance metrics used for each task are shown in Table~\ref{table:loss-and-metrics}.
\begin{equation}\label{eqn:evaluation-definition}
    \mathbb{E}_{I, a_{x} \sim D_{\text{test}}}\; \mathcal{M}_{x} (T^{\;y \to x} \vec{v}_{y}, I, a_{x}) \;\; \text{s.t.} \;\; \vec{v}_{y} = \arg\min_{\vec{v}} \; \mathbb{E}_{I, a_{y} \sim D_{\text{train}}}\; \mathcal{L}_{y} (\vec{v}, I, a_{y})
\end{equation}

\begin{table}[t]
    \centering
    \begin{tabular}{l|ll}
        \toprule
        \textbf{Task} & \textbf{Loss Function} & \textbf{Performance Metric} \\
        \midrule
        Generation & $\mathbb{E}_{I \sim D_{\text{train}}} \; \| \epsilon - \epsilon_{\theta} ( \sqrt{\alpha_t} I + \sqrt{1 - \alpha_t} \epsilon, t, \vec{v} ) \|^2$ & $\mathbb{E}_{I \sim p_{\theta}(\cdot | \vec{v})} \; \mathds{1} [ \; \text{$I$ has the concept} \; ]$ \\[0.2cm]
        Detection & $\mathbb{E}_{I,b,w \sim D_{\text{train}}} \; [ w \cdot ( \vec{e}_{\text{object}}(I, b)^{T} \vec{e}_{\text{text}}(\vec{v})) ]$ & Mean Average Precision \\[0.2cm]
        Classification & $\mathbb{E}_{I,w \sim D_{\text{train}}} \; [ w \cdot (\vec{e}_{\text{image}}(I)^{T}  \vec{e}_{\text{text}}(\vec{v})) ]$ & Classifier Accuracy \\
        \bottomrule
    \end{tabular}
    \vspace{0.2cm}
    \caption{\small Loss Functions and Performance Metrics. We benchmark transfer of word embeddings for visual concepts across generation, detection, and classification. In each row, $I$ corresponds to an image, $b$ to an object bounding box, and $w \in \{-1, 1\}$ to a weight multiplied onto the loss function. This weight controls whether the objective is maximized or minimized, where $w = -1$ when the image and bounding box contain the target concept, and $w = 1$ otherwise. The functions $\vec{e}$ are image and text encoders that return vector representations: $\vec{e}_{\text{image}}$ is the CLIP vision encoder, $\vec{e}_{\text{text}}$ is the CLIP text encoder, and $\vec{e}_{\text{object}}$ is the OWL-v2 region feature proposer.}
    \vspace{-0.2cm}
    \label{table:loss-and-metrics}
\end{table}

In the context of text-to-image generation with Stable Diffusion 2.1 \cite{stable-diffusion}, we use the denoising loss function originally proposed in Ho et al. 2020 \cite{ddpm}, where the goal is to predict a noise map $\epsilon$ added to an image $I$ at a particular timestep in the diffusion process $t$. We optimize the word vector embedding $\vec{v}$ so that Stable Diffusion generates images of a particular class (such as black Labrador). This optimization uses a training dataset $D_{\text{train}}$, and a separate dataset $D_{\text{test}}$ that contains different images of the same visual concept (such as black Labrador) is used for evaluation. For evaluating generation, we measure the probability that generations contain the target visual concept, measured by OpenAI's pretrained CLIP L-14 model given the prompt "a photo of \{\texttt{visual\_concept\_name}\}". This procedure is Textual Inversion \cite{textual-inversion} with an additional transfer step, shown in Figure~\ref{fig:transfer-generation}.

We use standard loss functions and performance metrics adapted from recent literature when training and evaluating words optimized for visual concepts. Each loss function and performance metric is discussed further in Section~\ref{sec:loss-functions}. Now equipped for training, evaluating, and transferring words across models, we can ask our motivating question: \textit{do models learn similar words for the same concepts?}

\section{Soft Prompts Are Model-Specific}
\label{sec:finding-one}

When optimizing words for visual prediction and generation tasks, the solutions found are typically model-specific. To understand the extent of the problem, we conduct a large-scale experiment, training 1200 new words for 40 visually distinct concepts in four standard datasets, across three state-of-the-art models in text-to-image generation, open-set object detection, and zero-shot classification. We generalize Textual Inversion~\cite{textual-inversion} to detection, and classification tasks, and optimize the embeddings for new words to minimize task-specific loss functions, described in Section~\ref{sec:loss-functions}. After new words have been trained to optimal performance on the \textit{training} task (i.e. generating a black Labrador), we evaluate on a different \textit{test} task (i.e. detecting black Labradors) using metrics discussed in Section~\ref{sec:metrics}.

\subsection{Dataset Preparation}\label{sec:dataset-preparation}

We employ the 2014 ImageNet detection dataset~\cite{imagenet}, the DreamBooth dataset~\cite{dreambooth}, COCO~\cite{ms-coco}, and PASCAL VOC~\cite{pascal-voc}. For each dataset, we select 10 concepts uniformly at random from the available classes to use for benchmarking, and select 8 images per concept from the training set. See Appendix~\ref{appendix:concepts} These cover a wide range of concepts likely to be encountered in the wild. For ImageNet, each image is annotated with an integer class label, and a set of bounding boxes that contain the target concept. For the DreamBooth dataset, bounding box labels are missing. To obtain bounding box labels, we ran a pretrained OWL-v2 on every image using the name of the subject as the prompt, and manually verified the labels as correct. For COCO and PASCAL VOC, class labels are not present, so we assign each image a class label equal to the class of the largest bounding box.

\subsection{Model Details}\label{sec:model-details}

We analyze three state-of-the-art models in text-to-image generation, open-set object detection, and zero-shot classification. Each model accepts a text-based prompt as input, containing the new word to be optimized (such as \texttt{<dog>} for the dog concept). For generation, we choose Stable Diffusion 2.1~\cite{stable-diffusion}, a latent diffusion-based generative model. For detection, we select OWL-v2 \cite{owl-vit}, a two-stage object detection model with a region proposal stage, and a classification stage that contrasts region features with text encodings of class names. For classification, we employ Data Filtering Networks~\cite{dfn}, which use CLIP-based~\cite{clip} contrastive training on a filtered dataset. These models have different input requirements. We resize images to 768x768 pixels when optimizing for Stable Diffusion \cite{stable-diffusion}, 960x960 for OWL-v2 \cite{owl-vit}, and 224x224 for Data Filtering Networks \cite{dfn}.

\subsection{Experiment Details}\label{sec:experiment-details}

\paragraph{Training} We take all combinations of models, datasets, and concepts, and perform 10 randomized trials, where we vary the initialization word used to seed the optimization algorithm. Initialization words are selected as the closest single token in the model's tokenizer to the name of a concept in the dataset. For example, 'sombrero' tokenizes to multiple subwords, so we use 'hat' for its initialization word. This choice ensures that our experiment accounts for both good and poor initializations. We optimize the embeddings for new word tokens using the Adam \cite{adam} optimizer with a learning rate of 0.0001, and a batch size of 8 (these hyperparameters are shared across all models). We train for 1000 gradient descent steps, and report final performance metrics using the optimized word embedding.

\begin{figure}[t]
    \centering
    \includegraphics[width=\linewidth]{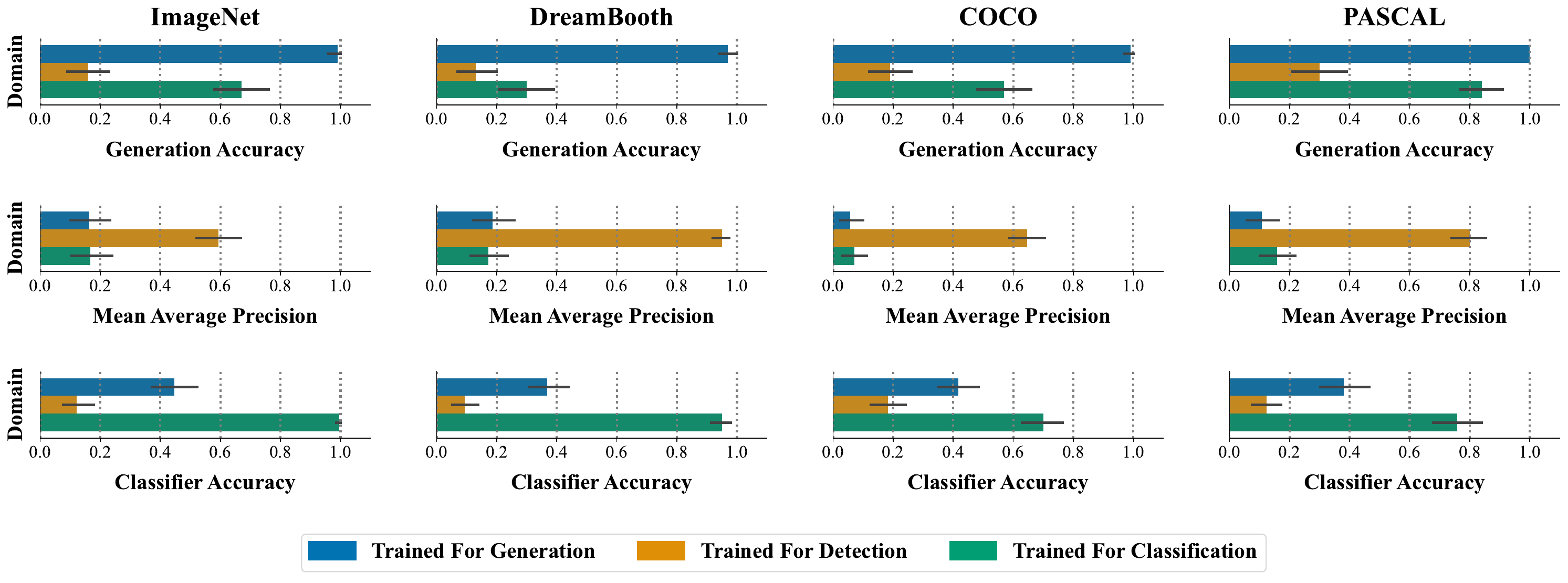}
    \caption{\small Visual word embeddings trained for one task (i.e. generation) perform well on that task, but may not perform well when transferred to another task (i.e. generation $\rightarrow$ detection). In certain directions, such as classification $\rightarrow$ generation, transfer works better than others. To understand when transfer fails, we perform extensive ablations across four standard datasets, and three models in generation, detection, and classification.}
    \vspace{-0.2cm}
    \label{fig:unconstrained}
\end{figure}

\begin{wrapfigure}[28]{r}{0.4\linewidth}
    \centering
    \vspace{-0.8cm}
    \includegraphics[width=\linewidth]{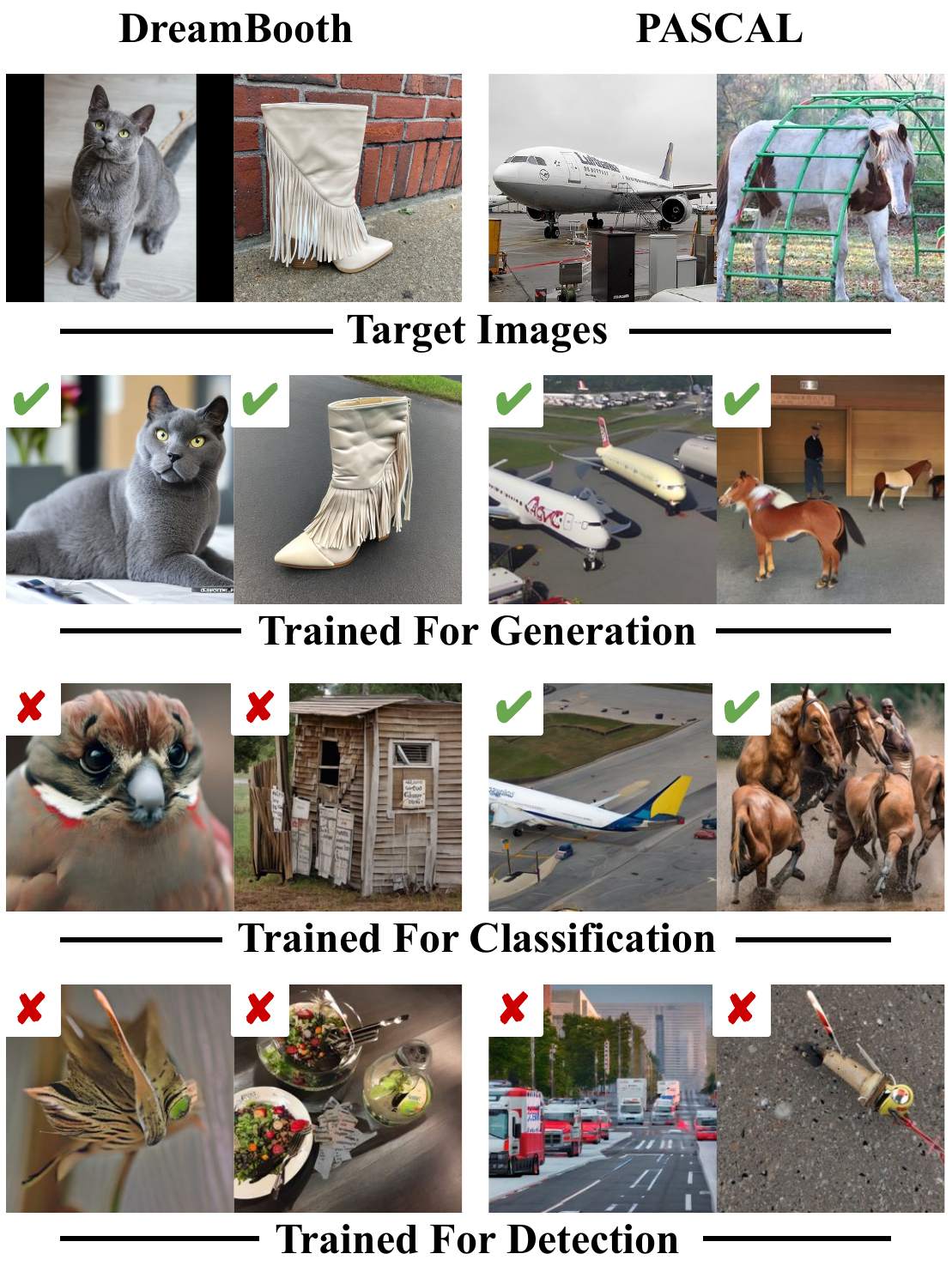}
    \caption{\small Generations (rows 2-4) from Stable Diffusion for target concepts (top row) from the DreamBooth and PASCAL datasets. The second row trains word embeddings for generation. The third row transfers word embeddings from classification to generation. The final row transfers from detection. Words trained for generation capture fine-grain details. Words trained for classification work for common concepts on PASCAL, but fail at fine-grain concepts on DreamBooth. Words trained for detection generally don't transfer.}
    \label{fig:case-study}
\end{wrapfigure}

\paragraph{Loss Functions}\label{sec:loss-functions} For generation, we employ the standard reparameterized denoising objective, introduced by Ho et al. in DDPM~\cite{ddpm}. For detection, we maximize the cosine similarity between the text and region feature containing the target object, and minimize cosine similarity to all other region features proposed by OWL-v2 \cite{owl-vit} in the image. For classification, we maximize cosine similarity between text and images of the target concept, and minimize cosine similarity to images that don't contain the target concept. Table~\ref{table:loss-and-metrics} shows the exact loss definitions.

\paragraph{Metrics}\label{sec:metrics} For generation, we report the rate at which an OpenAI CLIP L-14 \cite{clip} classifier predicts that generations are the target class (the set of class labels is the set of concepts names for that dataset from Appendix~\ref{appendix:concepts}), which we call Generation Accuracy in Table~\ref{table:loss-and-metrics}. For detection, we report the Mean Average Precision of bounding box predictions from OWL-v2 \cite{owl-vit} on images from held-out validation sets, annotated with bounding boxes. For classification, we report DFN CLIP-based~\cite{dfn,clip} classifier accuracy given images of the target concept, and unrelated concepts, from validation sets. All metrics are reported as 95\% confidence intervals over 100 randomized trials.

\subsection{Are Words Transferable?}
\label{sec:finding-one-conclusion}

Results of the experiment in Figure~\ref{fig:unconstrained} show that words optimized for visual tasks can perform great in-domain, but are typically not re-usable. In most transfer scenarios, words optimized for one task don't solve a different task than they were trained on with comparable fidelity to in-domain training. Words optimized for classification transfer best, achieving up to 84\% of the performance of in-domain training for generation (PASCAL), and up to 28\% of the in-domain detection performance (ImageNet). Words optimized for detection are least transferable, achieving up to 26\% of in-domain classification performance (COCO), and up to 30\% of in-domain generation performance (PASCAL). Words optimized for generation are in the middle in terms of their transferability, attaining up to 59\% of the in-domain classification performance (COCO), and up to 28\% of the in-domain detection performance (ImageNet). Generation shows a significant difference in performance between words transferred from classification vs. detection, what's happening here?

\paragraph{Understanding The Results} Using generation as a case study, we show images generated by Stable Diffusion 2.1 in Figure~\ref{fig:case-study} using word embeddings trained for generation (second row), transferred from classification (third row), and from detection (fourth row). We select two fine-grain concepts from the DreamBooth dataset, and two common concepts from the PASCAL dataset. Word embeddings trained for generation succeed at learning both fine-grain details for subjects in the DreamBooth dataset, and common classes in PASCAL. For word embeddings trained for classification, however, fine-grain details are missed, but common classes are learned. Results trained for detection miss fine-grain details, and common classes when transferred to generation, explaining trends in Figure~\ref{fig:unconstrained}.

\begin{figure}
    \centering
    \includegraphics[width=\linewidth]{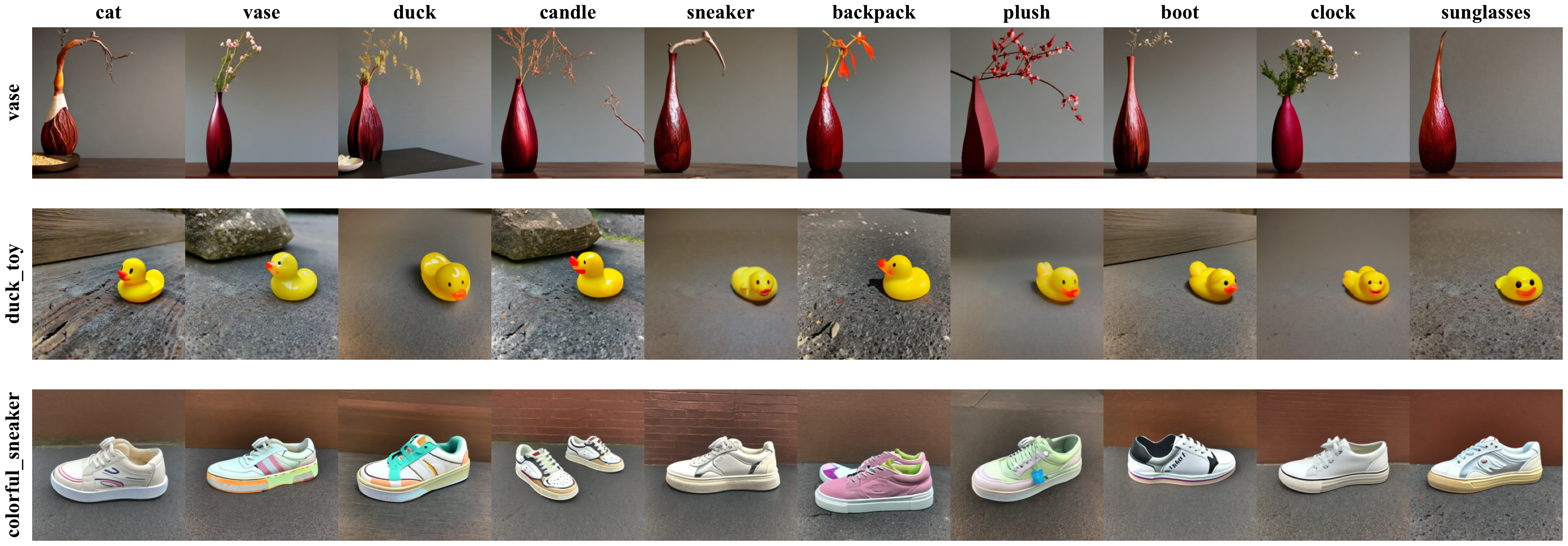}
    
    \vspace{0.5cm}
    
    \includegraphics[width=\linewidth]{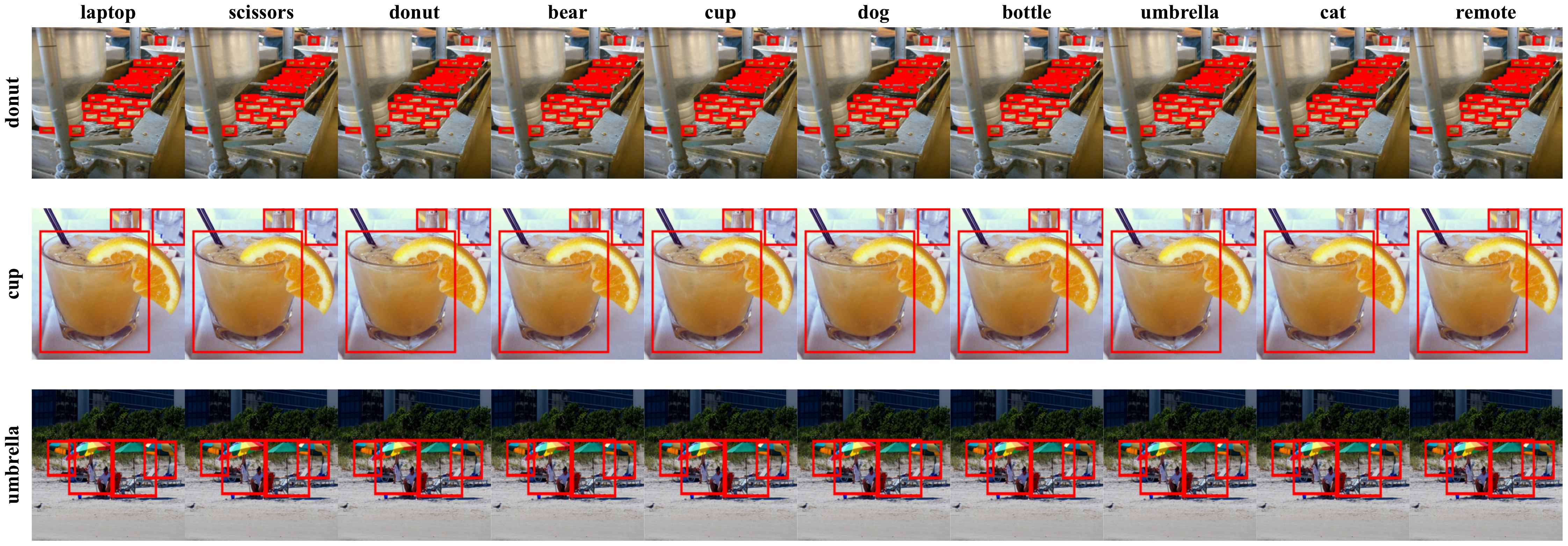}
    \caption{\small Example generations and detections for various concepts (row labels) using solutions found in the immediate neighborhood of unrelated words (column labels). We consistently find new words for generating and detecting arbitrary concepts near unrelated anchor words across ImageNet (examples in Appendix~\ref{appendix:more-examples}), DreamBooth (first three rows), COCO (last three rows), and PASCAL VOC (Appendix~\ref{appendix:more-examples}) datasets. The same objects are detected, and in several cases, near-identical images are generated.}
    \label{fig:fracture-examples}
\end{figure}

\section{Soft Prompts Are Fractured}
\label{sec:finding-two}

Results in Section~\ref{sec:finding-one} show that certain concepts transfer between certain models (discussed in Section~\ref{sec:finding-one-conclusion}), but most embeddings become random when transferred. We explore this phenomenon by considering a constrained objective for soft prompts in Equation~\ref{eqn:constrained-evaluation-definition}, where given an anchor word $w_{\text{anchor}}$ that we initialize $\vec{v}$ to, and a threshold $\delta$, we constrain solutions for $\vec{v}$ to an l2-ball of radius $\delta$, implemented using projected gradient descent. Transfer and evaluation remain the same as discussed in Section~\ref{sec:experiment-details}. We conduct a large-scale experiment, optimizing 4,800 words for 40 visual concepts across four standard datasets, three models, and four constraint thresholds $\delta \in \{0.1, 0.2, 0.5, 1.0\}$.
\begin{equation}\label{eqn:constrained-evaluation-definition}
    \vec{v}_{y} = \arg\min_{\vec{v}} \; \mathbb{E}_{I, a_{y} \sim D_{\text{train}}}\; \mathcal{L}_{y} (\vec{v}, I, a_{y}) \;\; \text{s.t.} \;\; \frac{\| \vec{v} - y(w_{\text{anchor}}) \|_{2}}{\min_{w \neq w_{\text{anchor}}} \| y(w) - y(w_{\text{anchor}}) \|_{2}} \leq \delta 
\end{equation}

This experiment controls where solutions are located in the embedding space, to help us understand the relationship between their location, and what gets transferred. Equipped with this tool, we can ask \textit{where performant solutions are located, and why some solutions transfer better than others}.

\begin{figure}
    \centering
    \includegraphics[width=\linewidth]{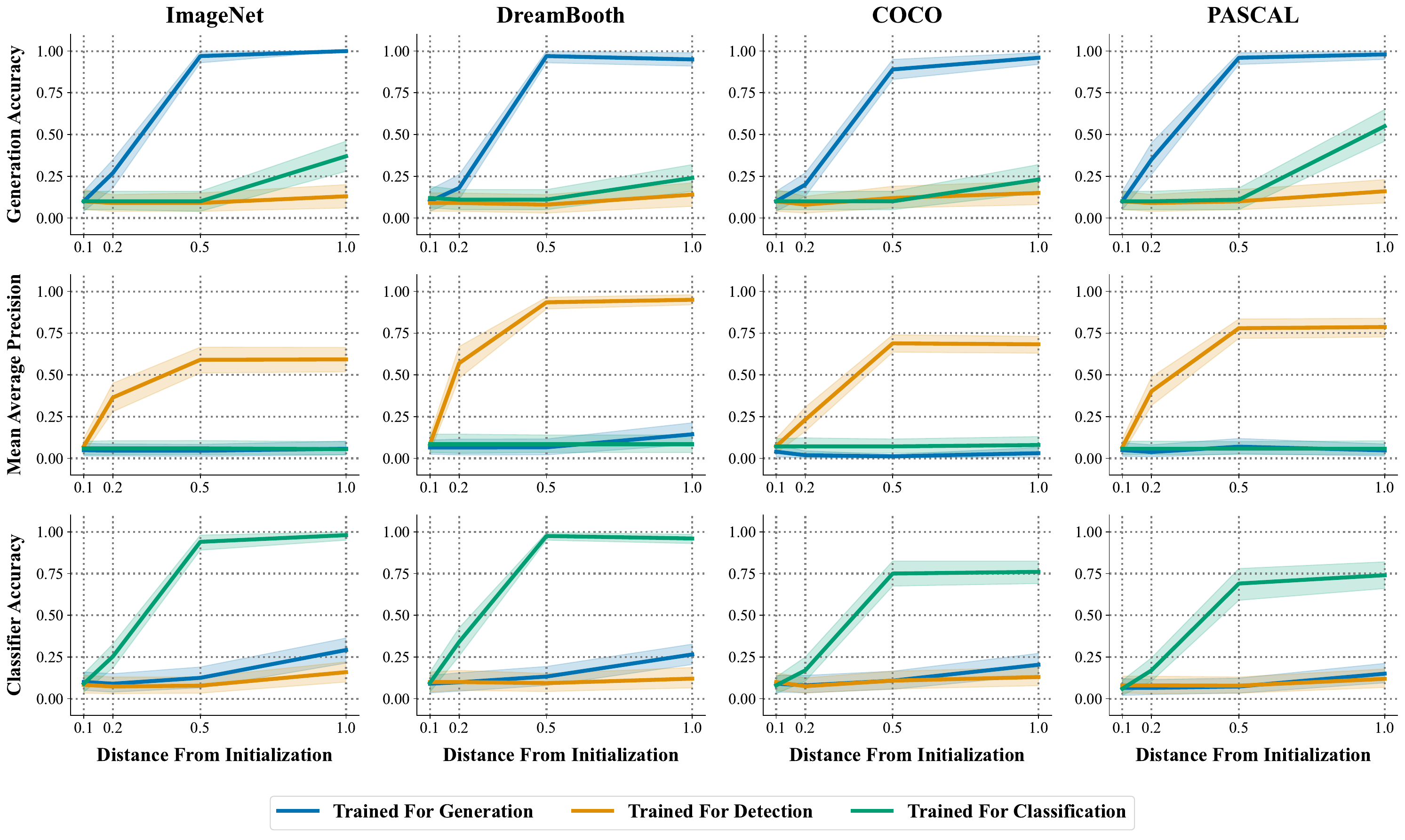}
    \caption{\small Performance (y-axis) of word vectors optimized to cause generation, detection, and classification of new visual concepts, for different constraint levels (x-axis). In-domain performance saturates at a constraint level of $\delta = 0.5$, which corresponds to solutions where the nearest existing word vector is the anchor $w_{\text{anchor}}$. Constrained solutions perform well in-domain, but typically don't perform well on transferred tasks for $\delta < 1$. Each line in the figure corresponds to the 95\% confidence interval of 100 randomized trials for 10 concepts, and 10 anchor words per dataset. Refer to Appendix~\ref{appendix:concepts} for the concepts and anchor words used for each dataset.}
    \vspace{-0.2cm}
    \label{fig:constrained}
\end{figure}

\subsection{Performant Solutions Are Everywhere}
\label{sec:solutions-everywhere}

Near the representation for any word in embedding space, there is a perturbation $\epsilon$ that causes models to generate, detect, and classify an arbitrary unrelated visual concept. For example, the representation in the top-left of Figure~\ref{fig:fracture-examples} is closest to the cat vector, but Stable Diffusion generates a red vase. This behavior is consistent across three tested models, four standard datasets, and 40 diverse visual concepts, suggesting it may be a general phenomenon in Large Multimodal Models. We name this phenomenon \textbf{fracturing} of the vector embedding space, as the set of word vectors that encode (i.e. generate) an arbitrary visual concept is disconnected, and parts of the set are close to every anchor word tested. Examples of these solutions are shown in Figure~\ref{fig:fracture-examples}, where each row corresponds to a visual concept from a standard dataset, and each column represents an anchor word. In several cases, an \textit{identical image} is generated by perturbations near two unrelated anchor words, such as generations for the duck concept (second row) for the vase (column two) and candle anchors (column four).

\paragraph{Performance Quickly Saturates} We measure performance of solutions in the fractured embedding space for different constraint levels $\delta \in \{0.1, 0.2, 0.5, 1.0\}$, and find their performance is indistinguishable from unconstrained solutions. Figure~\ref{fig:constrained} shows that performance saturates at a constraint level of $\delta = 0.5$, when the nearest neighbor is still the anchor word $w_{\text{anchor}}$. We observe that for all constraint levels $\delta > 1$, in-domain performance does not improve, despite the larger set of possible solutions. These results suggest that initialization is not very important, as performant solutions are likely close to any initialization. Instead, the data provided to the optimizer is likely more important.

\subsection{Perturbations Target The Final Layers}
\label{sec:activation-mismatch}

\begin{figure}
    \centering
    \includegraphics[width=\linewidth]{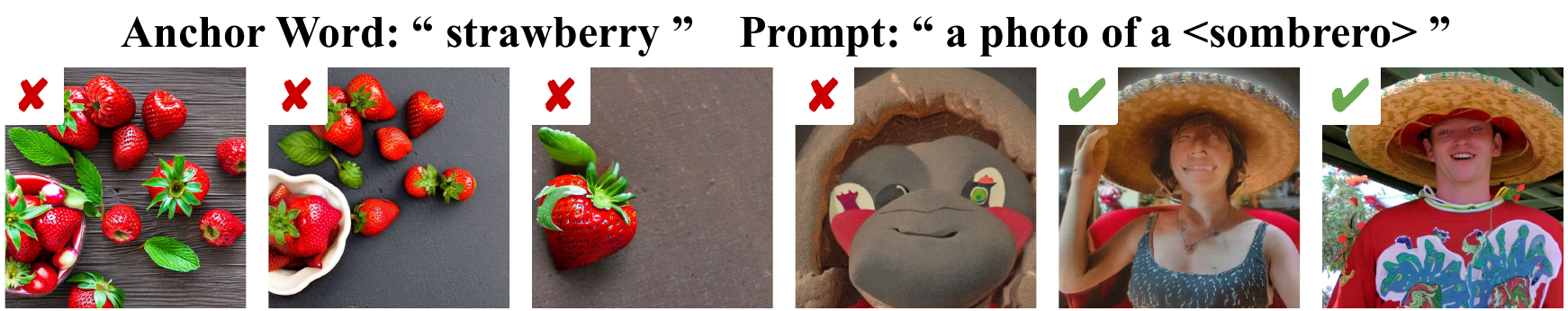}
    \includegraphics[width=\linewidth]{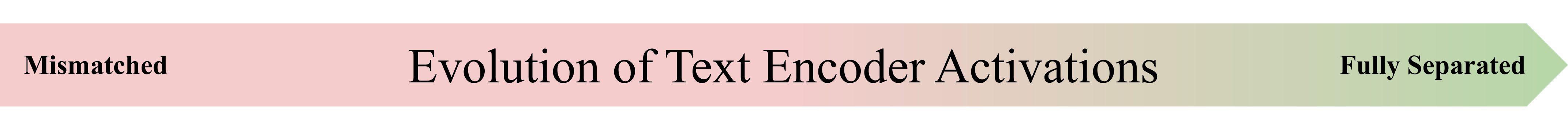}
    \includegraphics[width=\linewidth, trim={1.0cm 0 0 0}, clip]{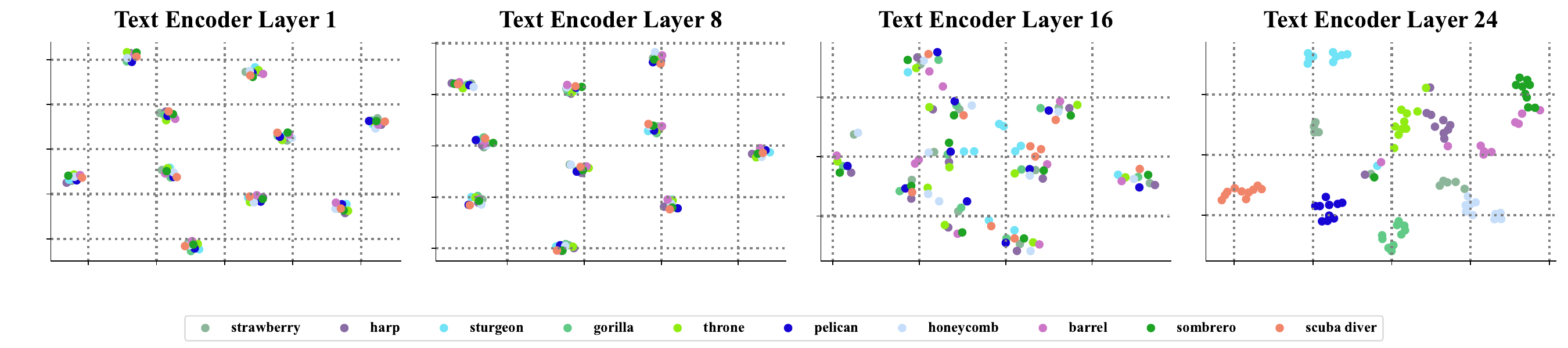}
    \caption{\small Prompts optimized for visual concepts target the final layers in text encoders. We show images generated by Stable Diffusion when truncating the text encoder to the first $N$ layers, and create TSNE visualizations of the text encoder activations for the pooling token at four evenly spaced layers. Each color represents a different visual concept. Clusters in plots 1-16 represent anchor words from Section~\ref{sec:solutions-everywhere}, which the activations cluster around instead of the target concept. When truncating the text encoder to just these layers, the anchor word (i.e. strawberry) is generated instead of the target concept (sombrero). Only by the final layers are clusters and generations correct. When transferred (visualizations in Appendix~\ref{appendix:more-visualizations}) clusters and generations stay mismatched.}
    \vspace{-0.2cm}
    \label{fig:tsne-visualization}
\end{figure}

Results in Section~\ref{sec:solutions-everywhere} show that performant solutions are located everywhere in the embedding space, and most of these solutions are non-transferable. How can we tell these solutions apart from the cases in Section~\ref{sec:finding-one-conclusion} that are transferable? One characteristic that identifies non-transferable solutions is their effect on the activations of the text encoder. Perturbative solutions like in Figure~\ref{fig:constrained} generally target the final layers of the text encoder, and lead to a disagreement between early and later layers. Figure~\ref{fig:tsne-visualization} shows generations from Stable Diffusion \cite{stable-diffusion} when truncating the text encoder to just the first $N$ transformer blocks (block = Norm $\to$ Attention $\to$ Residual $\to$ Norm $\to$ MLP $\to$ Residual). The bottom row shows TSNE visualizations of the pooling token activations at four evenly spaced layers in the text encoder of Stable Diffusion when generating concepts from the ImageNet~\citep{imagenet} task. Activations initially cluster around the anchor concept, i.e. strawberry, and generations from early layers yield the anchor concept, strawberry, instead of the target concept we optimized for, sombrero. When transferred (visualizations in Appendix~\ref{appendix:more-visualizations}), clusters and generations stay mismatched.

\section{Discussion}
\label{sec:discussion}

This work contributes a large-scale study of word embeddings that encode specific visual concepts across generation, detection, and classification tasks. We provide a benchmark for training soft prompts on a diverse set of visual concepts, and evaluating their transferability across three models. We show that certain embeddings are transferable between certain models, such as common concepts on the PASCAL task that transfer from classification $\to$ detection. In the majority of cases, soft prompts for visual concepts are model-specific, and to understand why, we conduct a large-scale ablation, training soft prompts constrained to the immediate neighborhood of different anchor words. We show that initialization does not matter as performant solutions are located everywhere in the embedding space, and non-transferable solutions resemble perturbations akin to adversarial examples.

Our work aims to galvanize the interoperability of large multimodal models following Figure~\ref{fig:teaser}, allowing prompts trained for generating black Labradors to be re-used for detection, and other tasks. Transferring prompts can significantly improve the adaptability and cost of machine learning systems by eliminating the need to re-train prompts when new models are released. We highlight the difficulty of transferring soft prompts for current multimodal models, and study why transfer often fails.

\clearpage

\bibliographystyle{abbrv}
\bibliography{main}

\clearpage
\appendix

\section{Limitations \& Safeguards}

We employ pretrained diffusion models, object detectors, and classifiers in this work, and these models are known to have biases, obtained from their training data. Diffusion models in-particular can generate harmful or dangerous content, including graphic imagery of violence, and pornography. We employ the Stable Diffusion safety checker to flag generations after transferring soft prompts for unsafe content as a mitigation strategy for this potential limitation. Transferring soft prompts currently does not perform very well outside of certain common concepts, and one limitation of this paper is its scope: we do not propose new methodology for transferring soft prompts with high fidelity. Rather, we benchmark popular methods for soft prompt-tuning on three recent models, and show that most prompts are not transferable. Our experiments suggest that non-transferable prompts have certain properties that can be used to identify them, but turning this identification strategy into a mitigation method is outside the scope of this paper, and a challenge that we leave for future research.

\section{Ethical Considerations}

Diffusion models currently require pristine data showing a subject in clear view in order to generate new photos of that subject. Transferring soft prompts from an object detector has the potential to allow for training on less pristine data that shows the subject amidst many distracting objects. One potentially harmful consequence of transfer between object detection models and generative models is related to privacy. Individuals that don't upload photos of themselves online are currently protected from their likeness being generated by diffusion models. However, transfer from object detectors to generative models would allow for their likeness to be generated, even when photos only show them in crowded spaces with many other people. Likewise, transferring prompts from generation to detection allows for the rapid creation of detectors for specific individuals. This technology could be used by malicious actors to track the activity of specific individuals, invading their privacy.

\section{Broader Impacts}

Transferring prompts for specialized tasks significantly improves the adaptability and cost of machine learning systems by removing the need to re-train when new models are released. The cadence of multimodal machine learning is such that new models are released every month, and the state-of-the-art is in constant flux. Currently, soft prompts trained for older models are discarded when newer models are released, or when the task changes (i.e. classification becomes detection). Enabling the re-use of soft prompts would allow users to download prompts trained by someone else, like plugins, even when the original use-case for that soft prompt was for a different task (such as generation).

One negative broader impact that results from improved transferability is that soft prompts encoding negative and harmful behaviours become easier to use and maintain. Currently, harmful prompts become obsolete quickly as newer models are released, but once they can be transferred, they become permanent. Mitigation strategies for this risk could involve moderating online databases containing soft prompts to remove ones that perpetuate harmful behaviors, and filtering the outputs of models using the soft prompts to directly remove the harmful content (in the same vein as a safety checker).

\section{Selected Concepts \& Anchor Words}\label{appendix:concepts}

In this section, we discuss the concepts that were selected from ImageNet \cite{imagenet}, COCO \cite{ms-coco}, PASCAL \cite{pascal-voc}, and the DreamBooth dataset \cite{dreambooth}. These concepts were selected uniformly at random without replacement from the available classes in each dataset. Ten classes were sampled per dataset in order to reduce the computational complexity of the experiments in the paper (results take 3 days to produce on just 40 visual concepts). These classes cover a diverse set of visual concepts. 

On the ImageNet dataset~\cite{imagenet}, we select \texttt{['strawberry', 'harp', 'sturgeon', 'gorilla', 'throne', 'pelican', 'honeycomb', 'barrel', 'sombrero', 'scuba diver']} as target concepts. 

On the DreamBooth Dataset~\cite{dreambooth}, we select \texttt{['cat2', 'vase', 'duck\_toy', 'candle', 'colorful\_sneaker', 'backpack\_dog', 'grey\_sloth\_plushie', 'fancy\_boot', 'clock', 'pink\_sunglasses']} as target concepts. 

On the COCO dataset~\cite{ms-coco}, we select \texttt{['laptop', 'scissors', 'donut', 'bear', 'cup', 'dog', 'bottle', 'umbrella', 'cat', 'remote']} as target concepts.

On the PASCAL VOC dataset~\cite{pascal-voc}, we select \texttt{['airplane', 'bicycle', 'bird', 'boat', 'person', 'train', 'car', 'cat', 'horse', 'cow']} as target concepts.

In addition to selecting concepts, we select anchor words that tokenize to a single token across all of the tested models. These are derived from the above target concepts. 

On the ImageNet dataset~\cite{imagenet}, we select \texttt{['strawberry', 'harp', 'sturgeon', 'gorilla', 'throne', 'pelican', 'honeycomb', 'barrel', 'hat', 'scuba']} as anchor words. 

On the DreamBooth Dataset~\cite{dreambooth}, we select \texttt{['cat', 'vase', 'duck', 'candle', 'sneaker', 'backpack', 'plush', 'boot', 'clock', 'sunglasses']} as anchor words. 

On the COCO dataset~\cite{ms-coco}, we select \texttt{['laptop', 'scissors', 'donut', 'bear', 'cup', 'dog', 'bottle', 'umbrella', 'cat', 'remote']} as anchor words.

On the PASCAL VOC dataset~\cite{pascal-voc}, we select \texttt{['airplane', 'bicycle', 'bird', 'boat', 'person', 'train', 'car', 'cat', 'horse', 'cow']} as anchor words.

\begin{table}[t]
    \centering
    \begin{tabular}{lr}
        \toprule
        \textbf{Hyperparameter Name} & \textbf{Hyperparameter Value} \\
        \midrule
        Generation Model Name & Stable Diffusion 2.1 \cite{stable-diffusion} \\
        Generation Model HuggingFace ID & \texttt{stabilityai/stable-diffusion-2-1} \\
        Generation Image Size & 768 x 768 \\
        Detection Model Name & OWL-v2 \cite{owl-vit} \\
        Detection Model HuggingFace ID & \texttt{google/owlv2-base-patch16-ensemble} \\
        Detection Image Size & 960 x 960 \\
        Classification Model Name & Data Filtering Networks \cite{dfn} \\
        Classification Model HuggingFace ID & \texttt{apple/DFN2B-CLIP-ViT-L-14} \\
        Classification Image Size & 224 x 224 \\
        Examples Per Concept & 8 \\
        Embedding Vectors Per Concept & 4 \\
        Denoising Steps & 50 \\
        Batch Size & 8 \\
        Learning Rate & 1e-04 \\
        Gradient Descent Steps & 1000 \\
        Optimizer & Adam \\
        Adam Beta1 & 0.9 \\
        Adam Beta2 & 0.999 \\
        Adam Epsilon & 1e-08 \\
        Weight Precision & float16 \\
        \bottomrule
    \end{tabular}
    \vspace{0.5cm}
    \caption{Hyperparameters used in the experiments of the paper. These parameters are held constant across all datasets and models. These choices are adapted from relevant prior work. }
    \vspace{-0.7cm}
    \label{tab:hyperparameters}
\end{table}

\section{Hyperparameters}
\label{appendix:hyperparameters}

In this section, we enumerate the hyperarameters used in the experiments in the paper. We choose hyperparameters agnostic to the model and task, so that results in the experiments are general, and not specific to the model. In Table~\ref{tab:hyperparameters} we note the HuggingFace model ID used, model configuration details, and hyperparameters from training, and evaluation.

\section{More Examples}
\label{appendix:more-examples}

In this section, we show more examples of generations from Stable Diffusion for perturbations to various unrelated anchor words in the embedding space. We show results for all combinations of 10 target concepts (row labels) and 10 anchor words (column labels) on ImageNet \cite{imagenet}, COCO \cite{ms-coco}, PASCAL \cite{pascal-voc}, and the DreamBooth dataset \cite{dreambooth}. In several cases, nearly identical images are generated by Stable Diffusion for perturbations near to different unrelated anchor words. 

\begin{figure}[t]
    \centering
    \includegraphics[width=\linewidth]{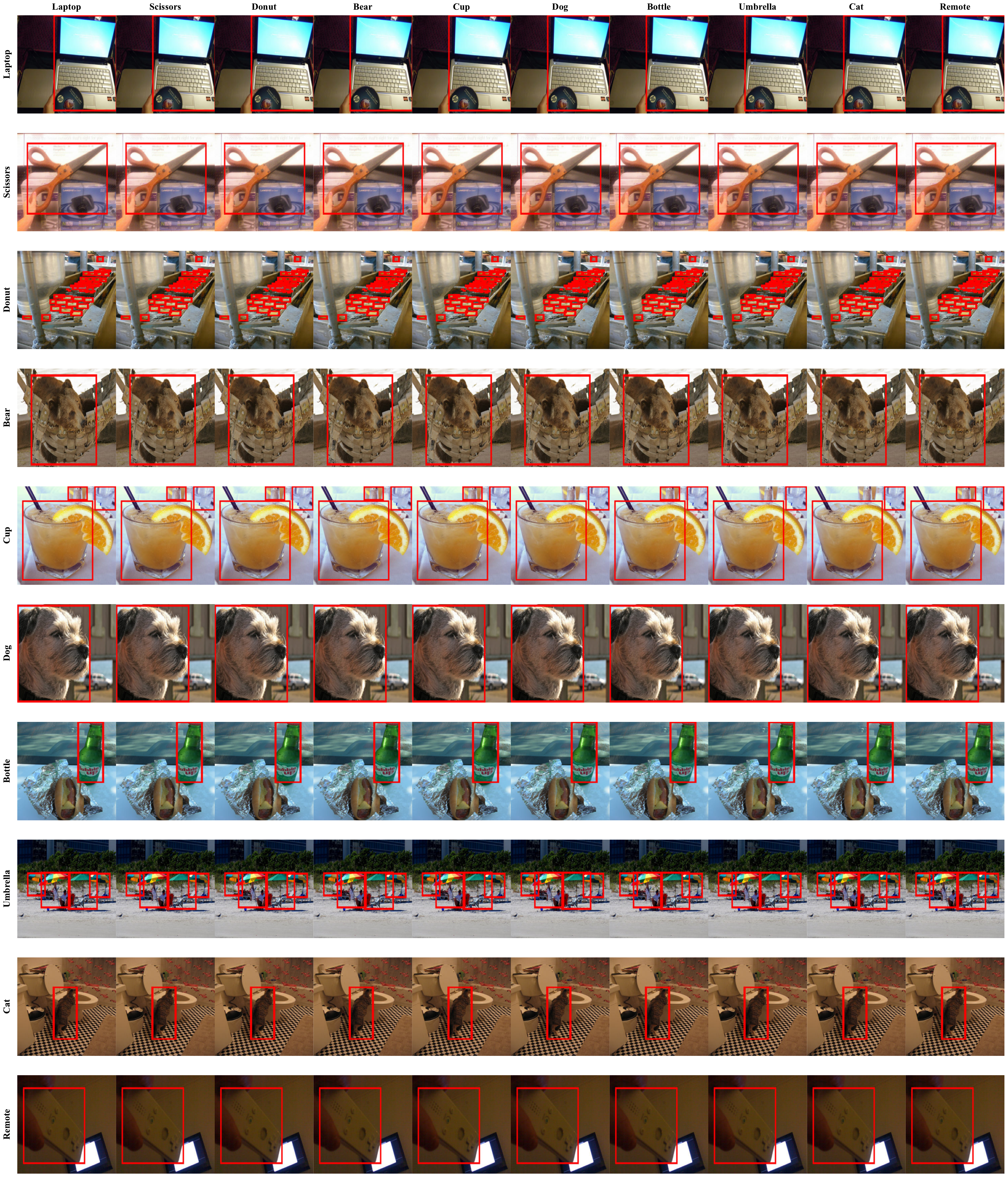}
    \caption{Visualizations of detections from OWL-v2 \cite{owl-vit} using new embeddings optimized for detecting visual concepts on COCO \cite{ms-coco}. Performant solutions for detecting arbitrary target concepts (row labels) are found with a constraint threshold $\delta = 0.5$ of unrelated anchor words (column labels). }
    \label{fig:more-coco-results}
\end{figure}

\begin{figure}[t]
    \centering
    \includegraphics[width=\linewidth]{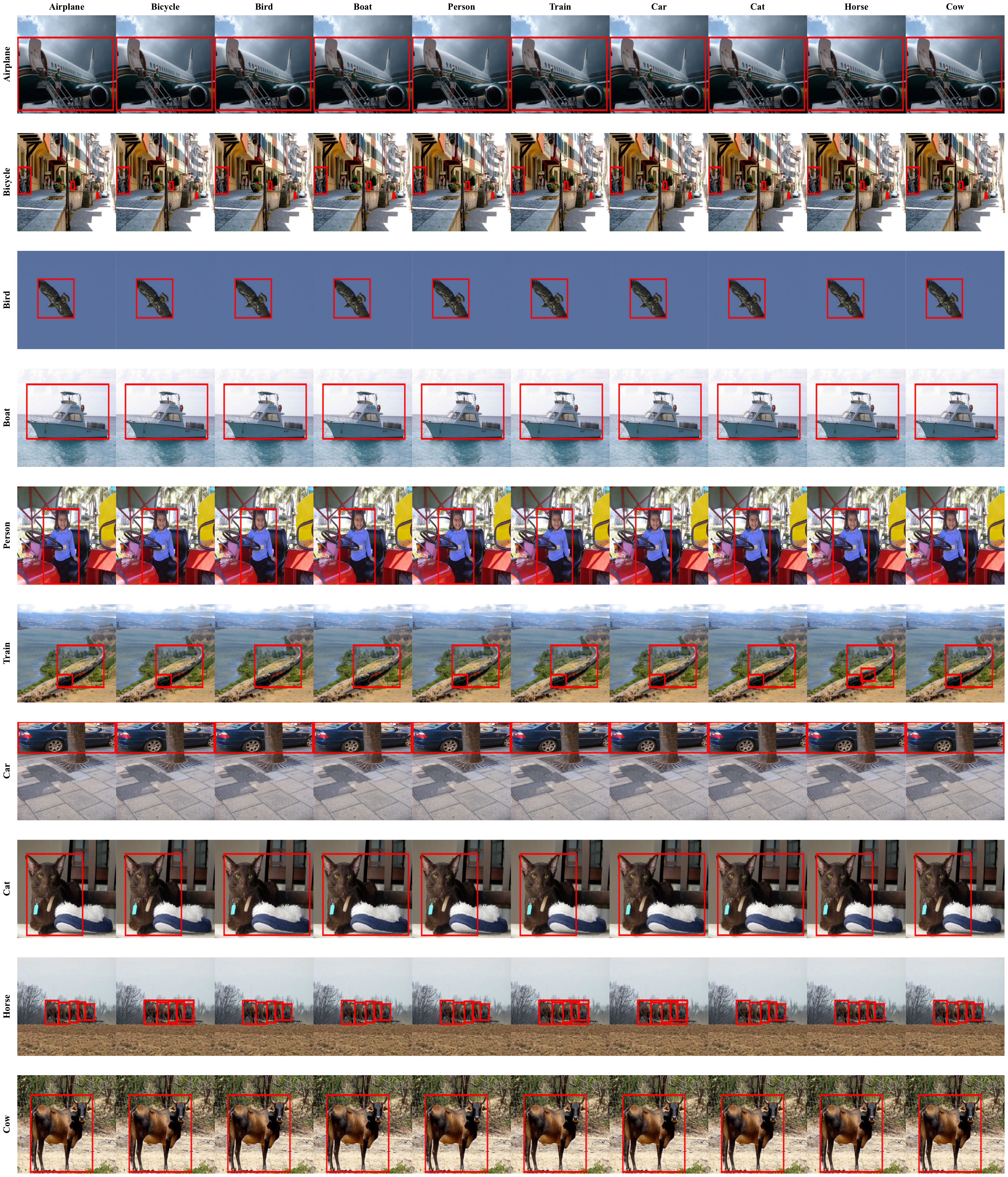}
    \caption{Visualizations of detections from OWL-v2 \cite{owl-vit} using new embeddings optimized for detecting visual concepts on PASCAL \cite{pascal-voc}. Performant solutions for detecting arbitrary target concepts (row labels) are found with a constraint threshold $\delta = 0.5$ of unrelated anchor words (column labels). }
    \label{fig:more-pascal-results}
\end{figure}

\begin{figure}[t]
    \centering
    \includegraphics[width=\linewidth]{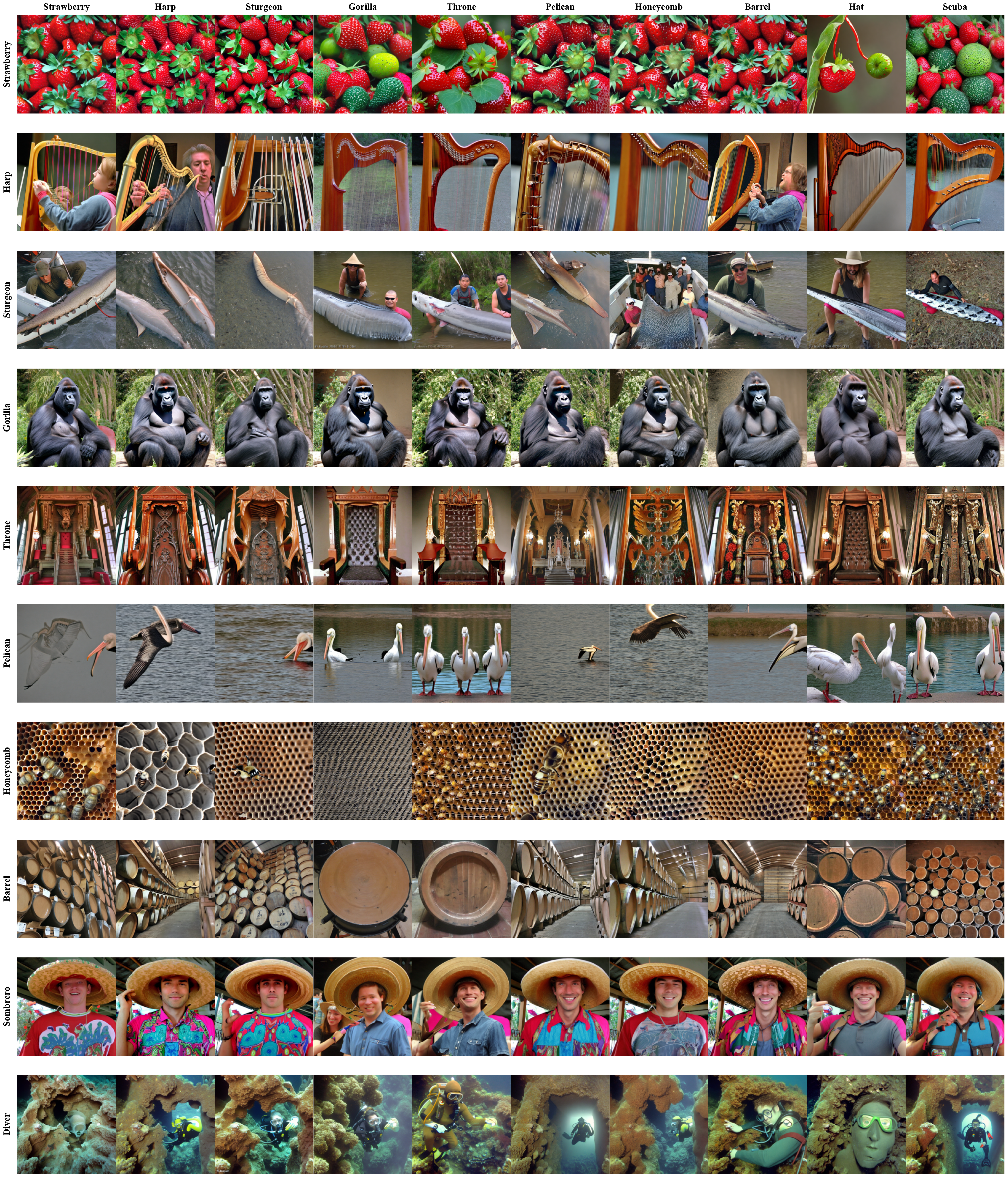}
    \caption{Visualizations of generations from Stable Diffusion 2.1 \cite{stable-diffusion} using new embeddings optimized for generating visual concepts on ImageNet \cite{imagenet}. Performant solutions for generating arbitrary target concepts (row labels) are found with a constraint threshold $\delta = 0.5$ of unrelated anchor words (column labels). In several cases, different solutions far apart generate the same image. }
    \label{fig:more-imagenet-results}
\end{figure}

\begin{figure}[t]
    \centering
    \includegraphics[width=\linewidth]{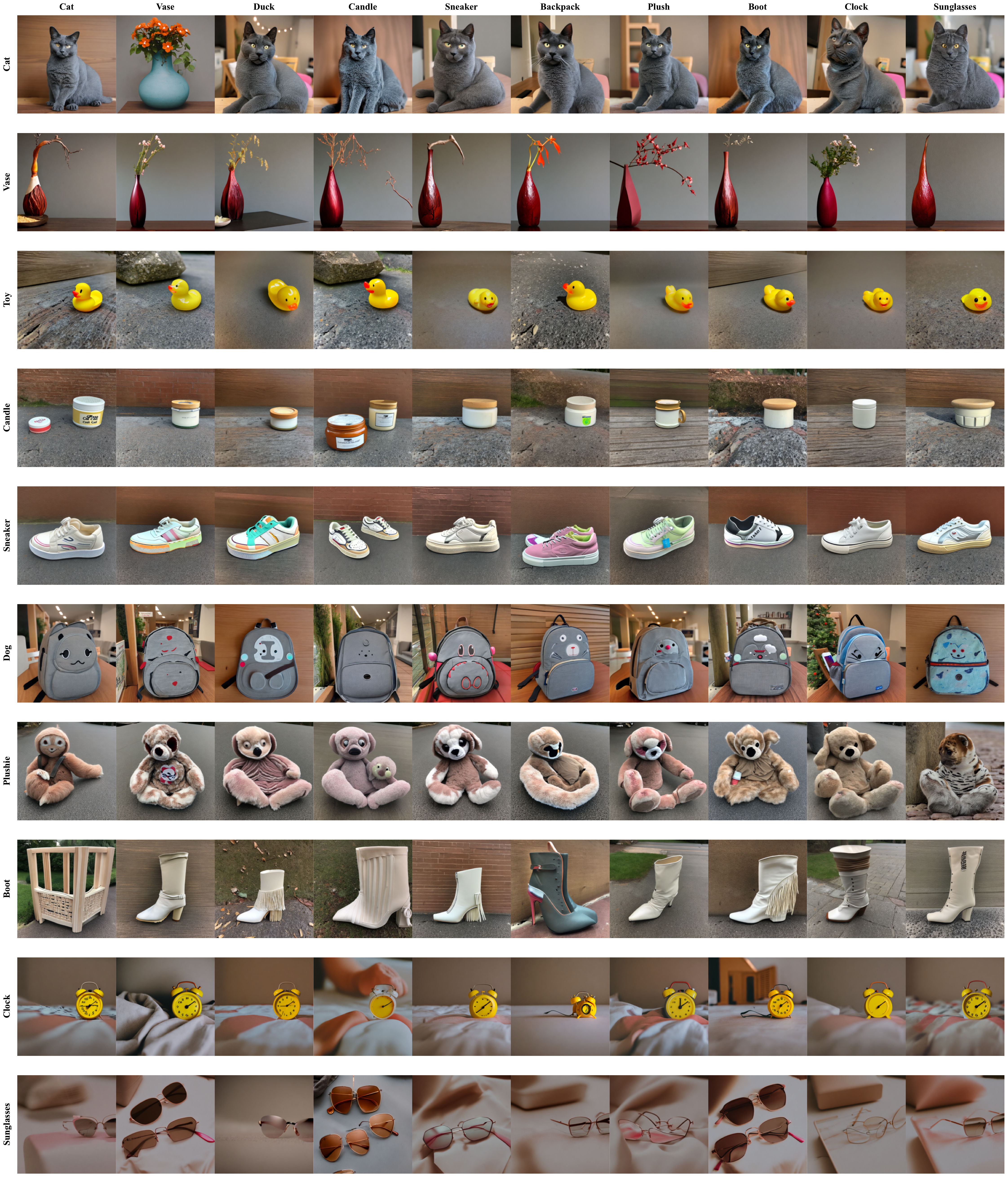}
    \caption{Visualizations of generations from Stable Diffusion 2.1 \cite{stable-diffusion} using new embeddings optimized for generating visual concepts on DreamBooth \cite{dreambooth}. Performant solutions for generating arbitrary target concepts (row labels) are found with a constraint threshold $\delta = 0.5$ of unrelated anchor words (column labels). In several cases, different solutions far apart generate the same image. }
    \label{fig:more-dreambooth-results}
\end{figure}

\section{More Visualizations}
\label{appendix:more-visualizations}

We provide more TSNE visualizations of the text encoder activations for different models and datasets in this section. Trends discussed in Section~\ref{sec:activation-mismatch} hold across all models and datasets. Perturbative soft prompts like those found in Section~\ref{sec:solutions-everywhere} target the final layers in text encoders, and early activations in text encoders disagree with later activations. Generating images when truncating the text encoder to the first $N$ layers leads to generations of the anchor work, instead of the target concept we are optimizing for (see Figure~\ref{fig:tsne-visualization}). When perturbative solutions are transferred, this transition stops.

\begin{figure}[t]
    \centering
    \includegraphics[width=\linewidth,trim={1cm 0 0 0},clip]{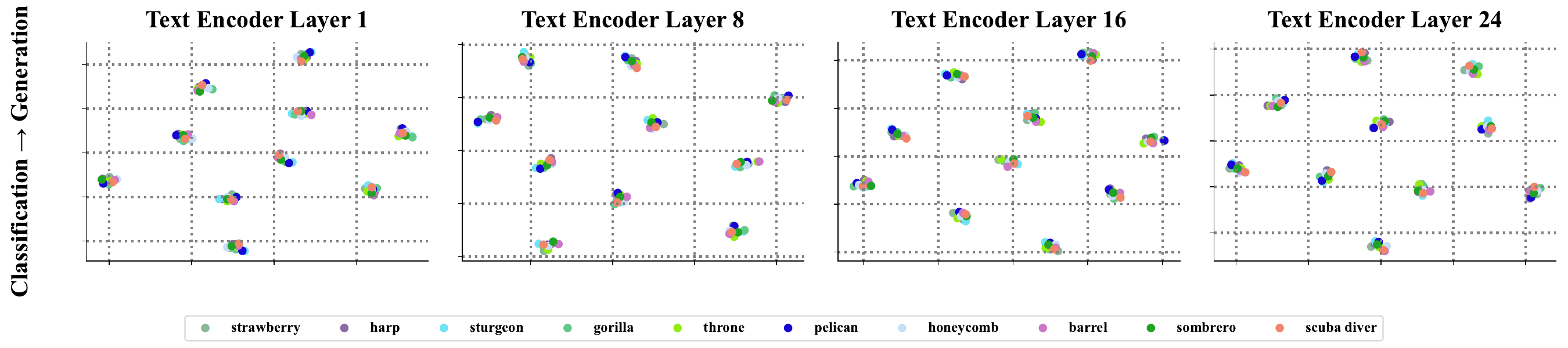}
    \caption{Visualizations of text encoder activations for Stable Diffusion 2.1 \cite{stable-diffusion} on ImageNet \cite{imagenet} at four evenly spaced layers when optimizing soft prompts for classifying visual concepts (colored points) and transferring to generation, constrained to the neighborhood of various anchor tokens (clusters in plots 1-24). The evolution of clusters towards clean separation for in-domain evaluation stops when soft prompts are transferred. Fine-tuning that targets the original model is lost.}
    \label{fig:imagenet-activations-transfer}
\end{figure}

Fine-tuning that targets the final layers of text encoders does not transfer, and Figure~\ref{fig:imagenet-activations-transfer} shows that activations stop clustering by concept (color) when soft prompts are transferred.

\begin{figure}[t]
    \centering
    \includegraphics[width=\linewidth]{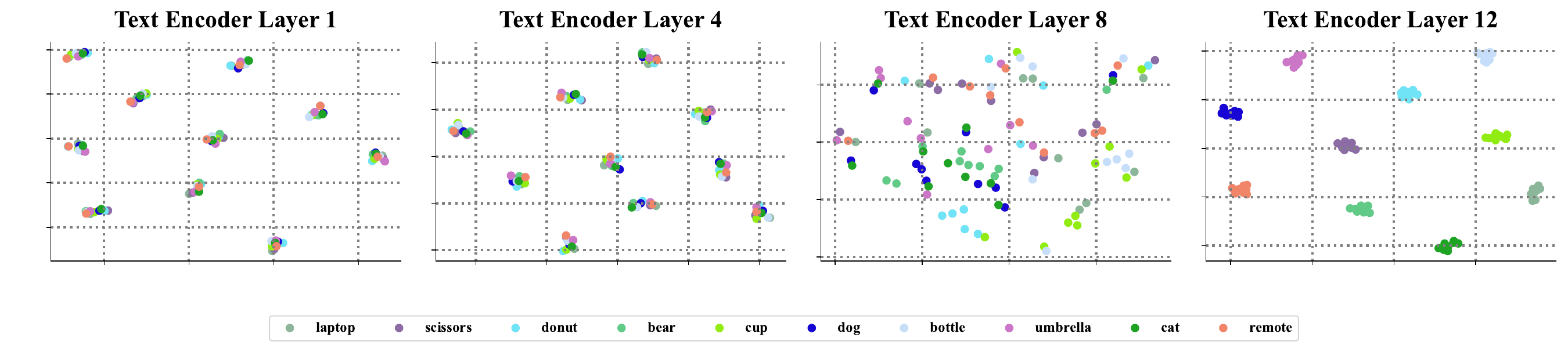}
    \caption{Visualizations of text encoder activations for OWL-v2 \cite{owl-vit} on COCO \cite{ms-coco} at four evenly spaced layers when optimizing soft prompts for detecting visual concepts (colored points), constrained to the neighborhood of various anchor tokens (clusters in plots 1-8).}
    \label{fig:coco-activations-detection}
\end{figure}

\begin{figure}[t]
    \centering
    \includegraphics[width=\linewidth]{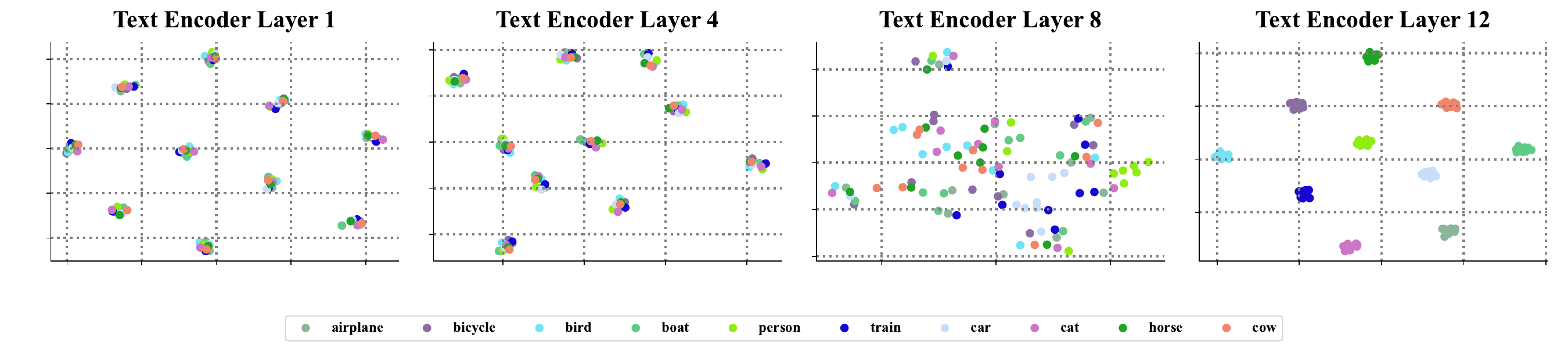}
    \caption{Visualizations of text encoder activations for OWL-v2 \cite{owl-vit} on PASCAL \cite{pascal-voc} at four evenly spaced layers when optimizing soft prompts for detecting visual concepts (colored points), constrained to the neighborhood of various anchor tokens (clusters in plots 1-8).}
    \label{fig:pascal-activations-detection}
\end{figure}

\begin{figure}[t]
    \centering
    \includegraphics[width=\linewidth]{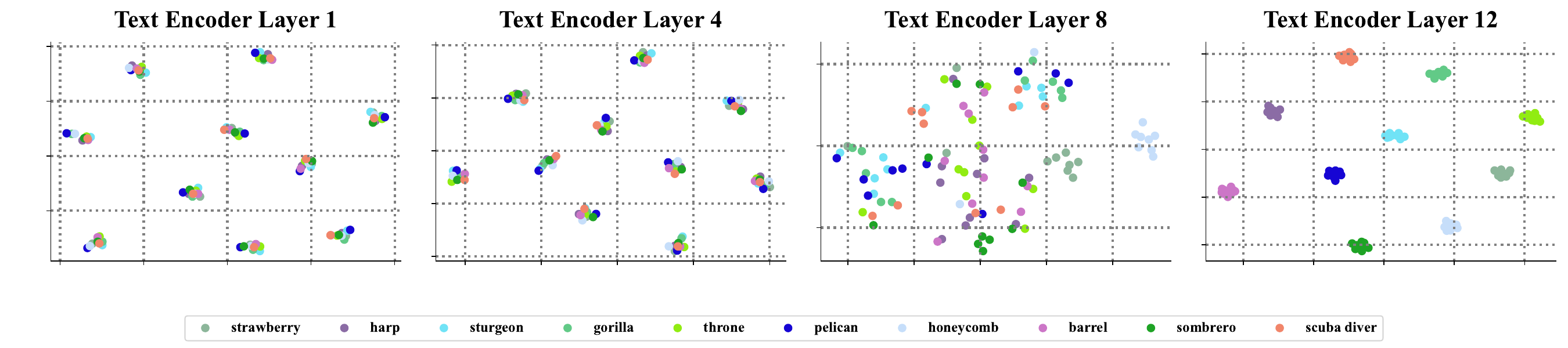}
    \caption{Visualizations of text encoder activations for DFN CLIP \cite{dfn, clip} on ImageNet \cite{imagenet} at four evenly spaced layers when optimizing soft prompts for classifying visual concepts (colored points), constrained to the neighborhood of various anchor tokens (clusters in plots 1-8).}
    \label{fig:imagenet-activations-classification}
\end{figure}

\begin{figure}[t]
    \centering
    \includegraphics[width=\linewidth]{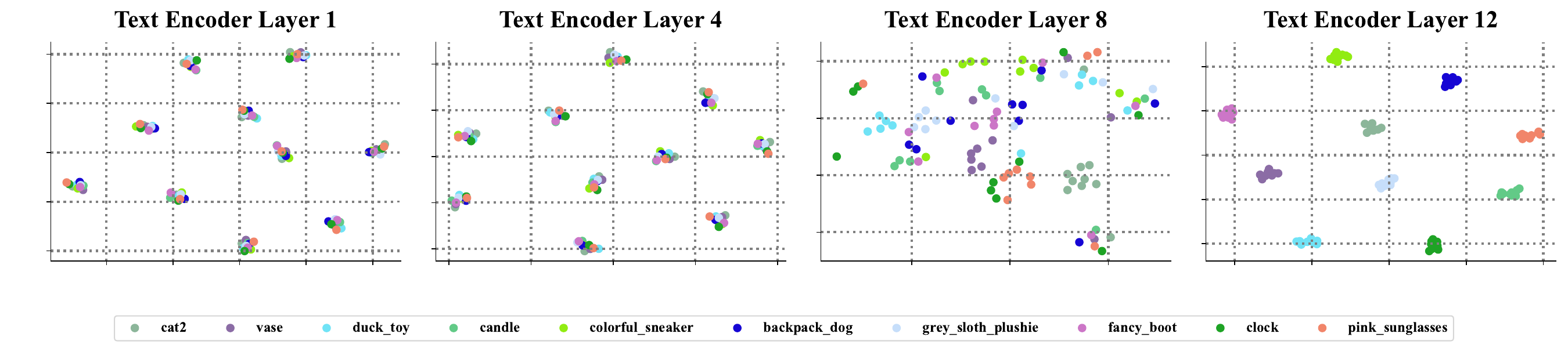}
    \caption{Visualizations of text encoder activations for DFN CLIP \cite{dfn, clip} on DreamBooth \cite{dreambooth} at four evenly spaced layers when optimizing soft prompts for classifying visual concepts (colored points), constrained to the neighborhood of various anchor tokens (clusters in plots 1-8).}
    \label{fig:dreambooth-activations-classification}
\end{figure}

\begin{figure}[t]
    \centering
    \includegraphics[width=\linewidth]{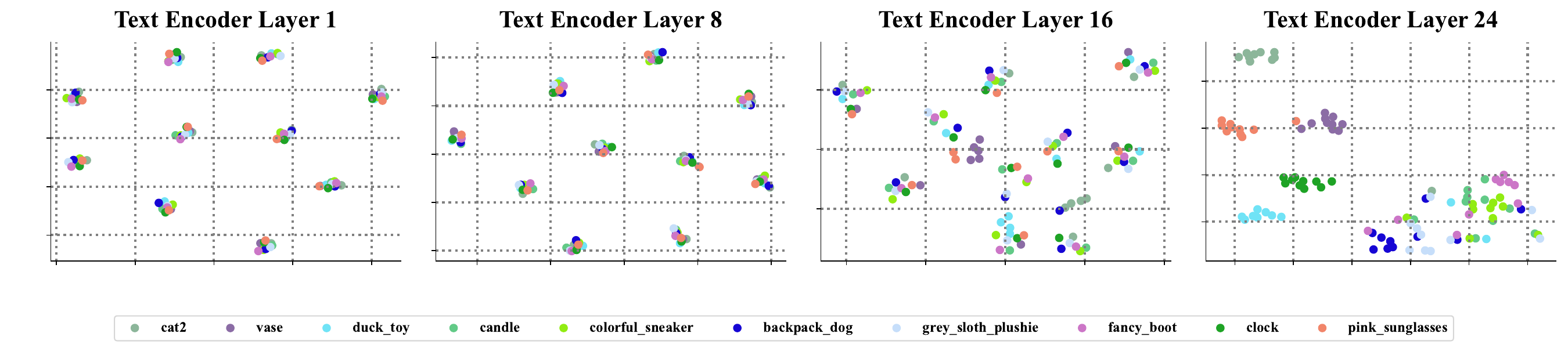}
    \caption{Visualizations of text encoder activations for Stable Diffusion 2.1 \cite{stable-diffusion} on DreamBooth \cite{dreambooth} at four evenly spaced layers when optimizing soft prompts for generating visual concepts (colored points), constrained to the neighborhood of various anchor tokens (clusters in plots 1-16).}
    \label{fig:dreambooth-activations-generation}
\end{figure}

\begin{figure}[t]
    \centering
    \includegraphics[width=\linewidth]{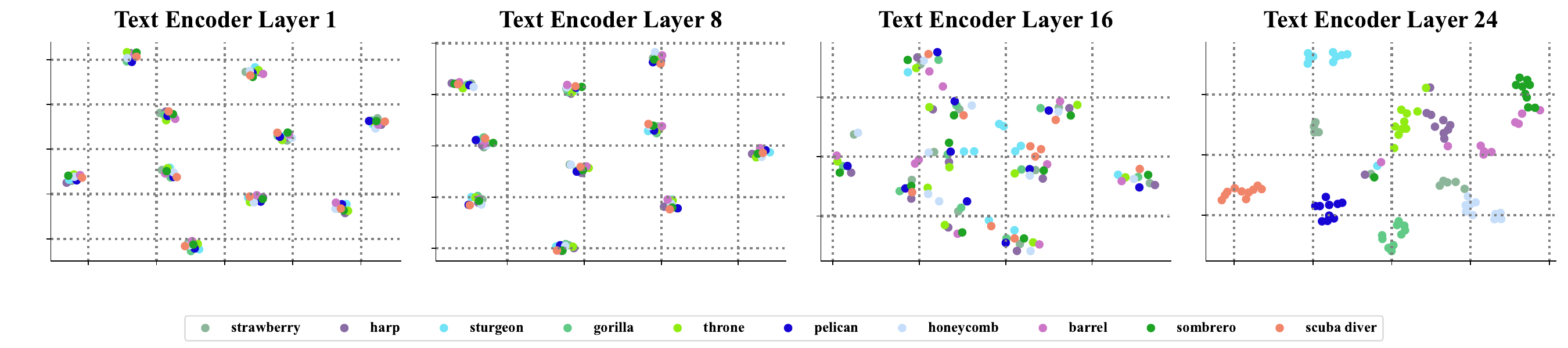}
    \caption{Visualizations of text encoder activations for Stable Diffusion 2.1 \cite{stable-diffusion} on ImageNet \cite{imagenet} at four evenly spaced layers when optimizing soft prompts for generating visual concepts (colored points), constrained to the neighborhood of various anchor tokens (clusters in plots 1-16).}
    \label{fig:imagenet-activations-generation}
\end{figure}

%\clearpage
%\input{sections/checklist}

\end{document}